\documentclass[]{cit_lab_mfr}

\usepackage{amsmath}
\usepackage{amssymb}
\usepackage{amsfonts}
\usepackage{nicefrac}

\usepackage[utf8]{inputenc}
\usepackage[T1]{fontenc}

\usepackage{xcolor}

\usepackage{booktabs}
\usepackage{multirow}
\usepackage{colortbl}
\usepackage{array}
\usepackage{adjustbox}

\usepackage{graphicx}
\usepackage{wrapfig}

\usepackage{enumitem}
\usepackage{pifont}

\usepackage{algorithm}
\usepackage{algpseudocode}

\usepackage{hyphenat}

\newcommand{\methodname}{SAVEMem}
\definecolor{isabelline}{rgb}{0.96, 0.94, 0.93}
\definecolor{lightblue}{rgb}{0.88, 0.93, 0.98}
\definecolor{grey}{rgb}{0.5, 0.5, 0.5}

\usepackage{listings}
\usepackage{xcolor}

\lstdefinestyle{mypython}{
    language=Python,
    basicstyle=\ttfamily\small,
    keywordstyle=\color{blue},
    stringstyle=\color{red!70!black},
    commentstyle=\color{green!50!black}\itshape,
    showstringspaces=false,
    breaklines=true,
    frame=single,
    numbers=left,
    numberstyle=\tiny\color{gray},
}

\title{\textbf{Semantic-Aware Adaptive Visual Memory for \\ Streaming Video Understanding}}

\author{%
\parbox{\textwidth}{\centering
Hang Wu$^{1}$ \qquad Sherin Mary Mathews$^{2}$  \qquad Yujun Cai$^{3, \S}$ \\[1mm]
\qquad Ming-Hsuan Yang $^{1}$ \qquad Yiwei Wang$^{1}$
}}

\affiliation{%
\parbox{\textwidth}{\centering\small
$^{1}$University of California, Merced \qquad $^{2}$US Bank \qquad  $^{3}$University of Queensland
}}

\contribution[\S]{Corresponding Author}

\renewcommand{\thefootnote}{\fnsymbol{footnote}}
\renewcommand{\thefootnote}{\arabic{footnote}}  % Reset for body

\abstract{
\begin{abstract}
/
Online streaming video understanding requires models to process continuous visual inputs and respond to user queries in real time, where the unbounded stream and unpredictable query timing turn memory management into a central challenge. Existing methods typically compress visual tokens via visual similarity heuristics, or augment compression with KV-cache-level retrieval. However, compression decisions rarely incorporate semantic signals, and retrieval is often added after compression is finalized, making the two stages hard to coordinate. We present \textbf{\methodname}, a training-free dual-stage framework that brings semantic awareness into memory generation and lets the retrieval scope adapt per query. In Stage~1, \methodname\ builds a three-tier streaming memory online under a constant memory budget. A fixed pseudo-question bank provides a lightweight semantic prior, so that long-term retention is shaped by semantic salience rather than visual similarity alone. In Stage~2, \methodname\ performs query-aware retrieval over this memory. An anchor-conditioned recency gate adapts the retrieval scope from short-term to mid- and long-term memory based on whether the query targets the present or the distant past. Within this scope, late interaction between query and memory tokens selects candidate frames for answering. Applied to Qwen2.5-VL without training, \methodname\ improves the OVO-Bench overall score from 52.27 to 62.69 and yields consistent gains on StreamingBench and ODV-Bench, while reducing peak GPU memory by 48\% at 128 frames over the backbone.
\end{abstract}
}

\date{\today}
\checkdata[Corresponding]{\url{hangwu@ucmerced.edu}, \url{vanora.caiyj@gmail.com}}
\checkdata[Code]{\url{https://github.com/wuhang03/savemem}}

\begin{document}
\maketitle

\begingroup
\renewcommand{\thefootnote}{\fnsymbol{footnote}}
% \footnotetext[3]{Work was done during their internship.}
\endgroup

\section{Introduction}
\begin{figure}
    \centering
    \includegraphics[width=1.0\linewidth]{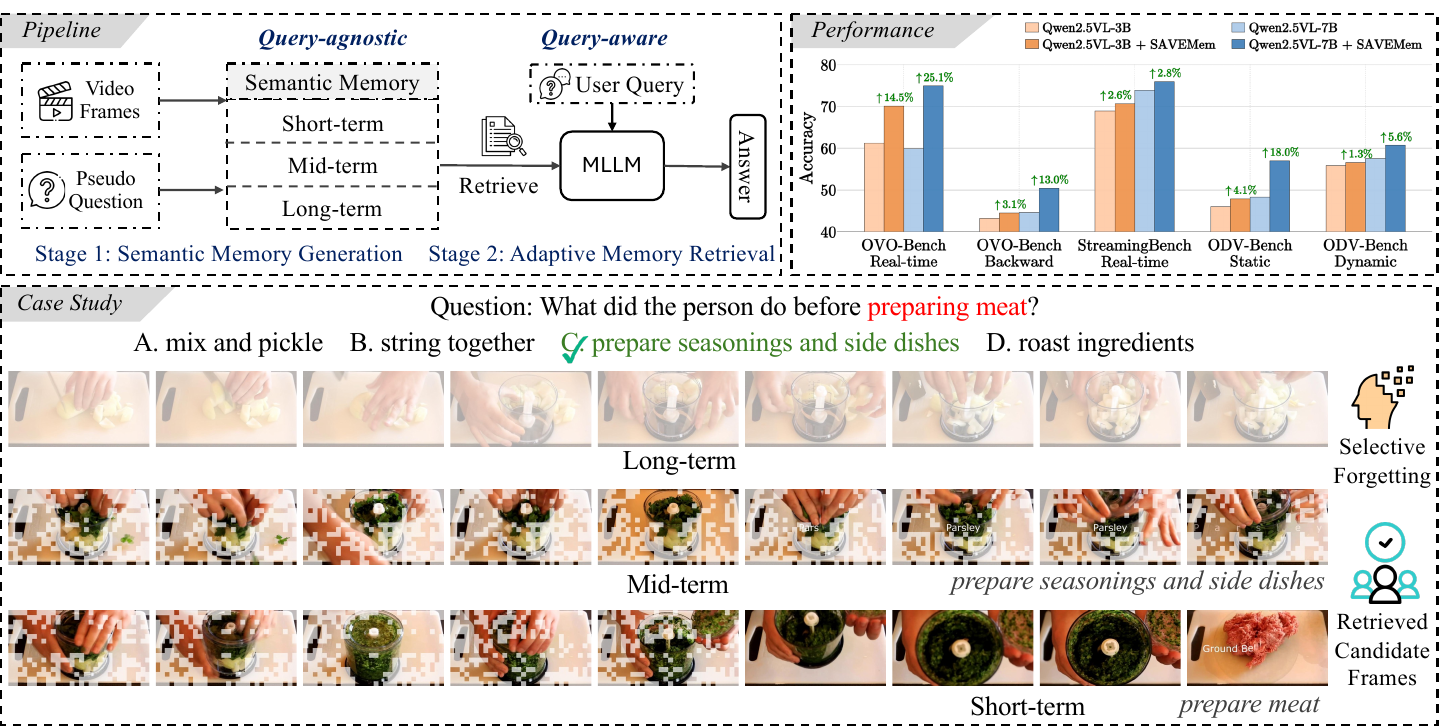}
    \caption{\textbf{Overview of \methodname.} A two-stage pipeline decouples query-agnostic semantic memory generation from query-aware retrieval, yielding consistent gains over Qwen2.5-VL backbones across multiple streaming benchmarks.}
    \label{fig:teaser}
\end{figure}

% task description for online streaming understanding: what is it, why is it important, what is the difficulty
% difficulty: causal constraints, long context and memory management
Online streaming video understanding requires models to process continuous visual inputs and respond to user queries in real time as the video unfolds, in contrast to offline settings where the entire video is available before inference~\cite{wang2023chatvideo,xu2025streamingvlm,zhang2025flash,wang2025streambridge}.  % what is it
This online setting is central to real-world applications such as robotic manipulation~\cite{liu2025aligning}, autonomous driving~\cite{chen2023e2esurvey}, and smart glasses~\cite{lee2018interaction}.  % why is it important
Compared with offline video question answering on a closed clip, streaming introduces coupled difficulties that reshape the role of memory. A causal constraint restricts the model to frames in $[0, t]$ at time $t$, ruling out bidirectional attention and global pooling~\cite{chen2024videollm,wu2024longvideobench}. Moreover, the stream is in principle unbounded while queries arrive at unpredictable moments and may target either the present or distant past~\cite{niu2025ovo,lin2024streamingbench}, so the model must continually decide what to retain at multiple temporal granularities to answer whenever a query arrives.

% existing methods for online streaming understanding
% what did they to to solve the problems, what defects do they have
To meet these requirements under a finite budget, existing work mainly focuses on compressing visual tokens before they enter the LLM~\cite{yao2025timechat,zeng2025streamforest,bolya2022tome,xie2026fluxmem} via visual similarity heuristics, with FluxMem~\cite{xie2026fluxmem} as a strong recent example using a hierarchical memory governed by adaptive Otsu-based thresholds~\cite{otsu1979threshold}. Another line of work manages the KV cache during the prefill stage to cope with unpredictable queries: ReKV~\cite{di2025rekv} and LiveVLM~\cite{ning2025livevlm} retrieve query-relevant KV-cache entries at inference, while WeaveTime~\cite{zhang2026weavetime} triggers coarse-to-fine recall via prediction uncertainty. SimpleStream~\cite{shen2026simple} further shows that feeding only the last $N$ frames to an off-the-shelf VLM is already competitive on many benchmarks, hinting at a perception-memory trade-off between present- and past-oriented queries. % what did they to to solve the problems
Despite this progress, two aspects remain under-explored. On the one hand, compression decisions rely mostly on visual similarity, leaving limited room for semantic signals to shape long-term retention. On the other hand, retrieval is typically added at the KV-cache level after compression is finalized and often requires fine-tuning, making retention and retrieval hard to coordinate as a single pipeline. % what defects do they have

% our method
% how does our method solve previous problems
Motivated by these observations, we present \textbf{\methodname}, a training-free dual-stage framework that brings semantic awareness into compression and lets the retrieval scope adapt per query. As shown in Fig.~\ref{fig:teaser}, \methodname\ builds a three-tier streaming memory online in Stage~1, where a fixed pseudo-question bank provides a lightweight semantic prior so that long-term retention is shaped by semantic salience rather than visual similarity alone, all under a constant memory budget. In Stage~2, \methodname\ performs query-aware retrieval over the memory built in Stage~1, where an anchor-conditioned recency gate adapts the retrieval scope from short-term to mid- and long-term memory based on whether the query targets the present or the distant past. Within this scope, late interaction between query and memory tokens yields frame-level relevance scores, based on which candidate frames are retrieved and fed to the model for answering.
% how does our method perform
Applied to Qwen2.5-VL-7B~\cite{bai2025qwen2_5vl}, \methodname\ improves the OVO-Bench~\cite{niu2025ovo} overall score from 52.27 to 62.69, surpassing recent training-free competitors such as FluxMem~\cite{xie2026fluxmem} and HERMES~\cite{zhang2026hermes}. Consistent improvements on StreamingBench~\cite{lin2024streamingbench} and ODV-Bench~\cite{zeng2025streamforest} further suggest that decoupling query-agnostic compression from query-aware retrieval is a practical direction for scaling to long-form streaming video without training. 

Our main contributions can be summarized as below:

\begin{itemize}
    \item We present \methodname, a training-free dual-stage framework that introduces semantic priors into streaming compression and adapts the retrieval scope per query for online streaming video understanding.
    
    \item Applied to Qwen2.5-VL without training, \methodname\ improves over the backbones on three streaming benchmarks and compares favorably with recent training-free methods.
    
    \item Efficiency analysis shows sub-linear token growth and nearly flat peak GPU memory with respect to video length, reducing memory usage by 48\% at 128 frames over the backbone.
\end{itemize}

\section{Related Work}
 
%\subsection{Multimodal Large Language Models}
\noindent \textbf{Multimodal Large Language Models.}
Recent advances in Multimodal Large Language Models (MLLMs)~\cite{li2024llava_ov,wang2023see} have broadened their application to video understanding. 
Typically, these models comprise a visual encoder for extracting frame-level representations, a modality projector to map visual features into the language space, and a Large Language Model (LLM) to generate contextual responses~\cite{damonlpsg2023videollama,Maaz2023VideoChatGPT,bai2025qwen2_5vl,li2024llava_ov,zhang2024llavavideo,tang2025video,wang2025internvideo}.
Recent work further pushes video MLLMs toward more adaptive and reasoning-oriented behavior~\cite{ge2025framemind,wu2026camreasoner}, while effectively delivering relevant context to the LLM remains an open challenge across modalities~\cite{mei2025survey,fu2025contextnav}.
However, these models are inherently designed for static and offline settings where the input is a pre-loaded full video rather than a continuous stream, and they fail to adapt to dynamic real-world scenarios where video frames are processed sequentially and require real-time, temporally coherent, or even proactive responses~\cite{lin2024streamingbench,niu2025ovo,huang2025online}.
 
%\subsection{Streaming Video Understanding}
\noindent \textbf{Streaming Video Understanding.}
Understanding video streams in real time requires sequential processing of incoming frames~\cite{lin2024streamingbench,niu2025ovo,huang2025online,qian2024videostreaming}, and existing approaches fall into two broad categories.
One line of work manages KV caches during inference. ReKV~\cite{di2025rekv} retrieves query-relevant entries from stored caches, LiveVLM~\cite{ning2025livevlm} separates short- and long-term memory with online retrieval, and StreamMem~\cite{yang2025streammem} maintains bounded memory through continuous compression.
Another line of work compresses visual tokens upstream of the LLM. ToMe~\cite{bolya2022tome} merges similar tokens, and FluxMem~\cite{xie2026fluxmem} introduces a hierarchical memory governed by adaptive thresholds~\cite{otsu1979threshold}.
Beyond streaming video, a parallel line of work in long-context language modeling explores gated and utility-aware memory consolidation, suggesting that selectively writing salient information into a bounded memory is more efficient than uniform updates~\cite{mei2026gated}.
Despite this progress, most streaming methods rely on visual similarity and fix compression prior to the query. Recent work begins to address this issue. WeaveTime~\cite{zhang2026weavetime} uses uncertainty to trigger coarse-to-fine recall, and SimpleStream~\cite{shen2026simple} shows that using only the last $N$ frames can match many methods, which exposes a perception and memory trade-off that cannot adapt per query.
These observations motivate \methodname, which injects semantic signals into query-agnostic compression and performs query-driven retrieval over compressed tokens with a training-free ColBERT-style mechanism~\cite{khattab2020colbert}, enabling per-query trade-off navigation.

\section{Method}

\subsection{Problem Formulation}

\paragraph{Task Definition.}
We study \textit{online streaming video question answering} (StreamingVQA), where a model observes a continuous video stream $\mathcal{V} = \{v_t\}_{t=1}^{\infty}$ and must answer a natural-language query $q$ issued at timestamp $T$ using only the observed prefix:
\begin{equation}
    a = f\!\left(q,\, \mathcal{V}_{[0,T]}\right), \quad \mathcal{V}_{[0,T]} = \{v_t\}_{t=1}^{T}.
\end{equation}
Following standard practice~\cite{lin2024streamingbench,niu2025ovo}, we adopt the \textit{pseudo-streaming} protocol: the video is truncated at $T$ and $q$ is withheld until after memory construction, correctly enforcing causal access and query-agnostic encoding without imposing frame-by-frame latency constraints (see Appendix~\ref{supp:problem}).

\paragraph{Streaming Constraints.}
Three constraints jointly define the streaming regime.
\textbf{(C1) No future access}: the model may not use any frame $v_t$ with $t > T$.
\textbf{(C2) Query-agnostic encoding}: all memory construction must proceed without conditioning on $q$; fixed surrogate signals such as chat template tokens~\cite{yang2025streammem} remain compliant as they carry no information about the actual query.
\textbf{(C3) Bounded memory}: total memory is capped at a constant $B$ independent of video length.
See Appendix~\ref{supp:problem} for more details.

\paragraph{Two-Phase Paradigm.}
Existing methods address C1--C3 via a two-phase design: \textit{encoding} builds a compact visual memory as frames arrive in a query-agnostic manner; \textit{inference}, triggered at $T$, retrieves relevant entries conditioned on $q$.
This is the standard structure in Flash-VStream~\cite{zhang2025flash}, ReKV~\cite{di2025rekv}, LiveVLM~\cite{ning2025livevlm}, StreamMem~\cite{yang2025streammem}, StreamForest~\cite{zeng2025streamforest}, and Fluxmem~\cite{xie2026fluxmem}.

\begin{figure}
    \centering
    \includegraphics[width=1.0\linewidth]{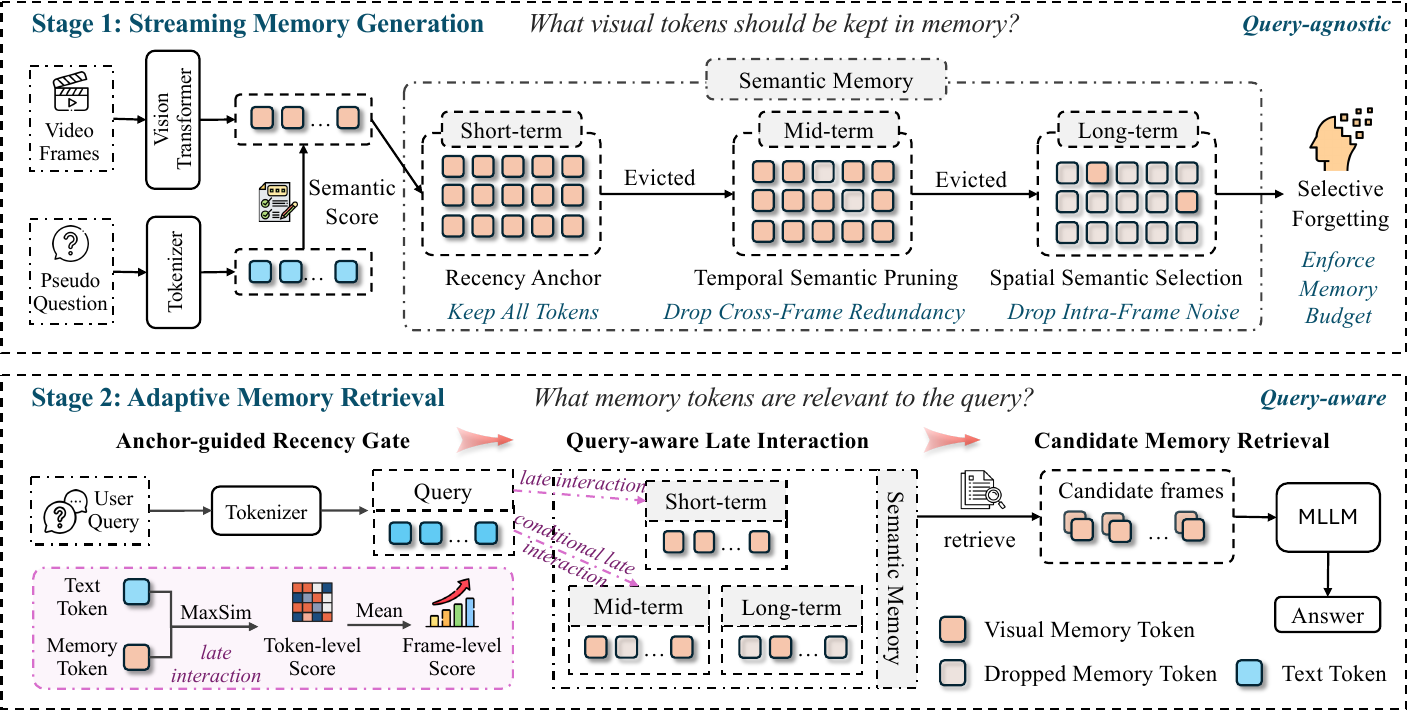}
    \caption{\textbf{Framework of \methodname.} Stage~1 builds a three-tier streaming memory query-agnostically through a Recency Anchor, Temporal Semantic Pruning, and Spatial Semantic Selection guided by a pseudo-question prior; Stage~2 retrieves query-relevant frames query-awarely via an anchor-conditioned recency gate and late interaction between query and memory tokens.}
    \label{fig:framework}
\end{figure}

%% ================================================================
\subsection{Overview}
\label{sec:overview}
\methodname\ is a training-free dual-stage framework for online streaming video QA that brings semantic awareness into compression and adaptive scope into retrieval. As shown in Fig.~\ref{fig:framework}, the pipeline is decoupled at query arrival time into a query-agnostic compression stage and a query-aware retrieval stage. Stage 1 builds a three-tier streaming memory online, where a fixed pseudo-question bank provides a lightweight semantic prior to guide cross-tier compression under a constant memory budget. Stage 2 then performs read-only retrieval over the aged tiers, where an anchor-conditioned recency gate adapts the scope to the query's temporal target and late interaction scores candidate frames for answering. The overall algorithm pipeline is shown in Algorithm~\ref{alg:methodname}.

\paragraph{Query-agnostic Semantic Prior.}
Visual similarity identifies redundant tokens but does not reflect semantic importance, since two tokens with similar neighbours may differ substantially in informational value. To inject a coarse semantic signal without access to the user query, we score each visual token $v$ against a fixed pseudo-question bank $Q$ via late-interaction MaxSim:
\begin{equation}
s(v) \;=\; \max_{q \in Q}\; \cos(v,\, q).
\label{eq:prior}
\end{equation}
The bank consists of a small set of generic probes covering visual semantics commonly queried in streaming video, including object presence, counting, action and event occurrence, scene change, and spatial layout. By taking the maximum similarity across probes, $s(v)$ captures whether a token aligns with at least one of these semantic axes, providing a coarse but query-agnostic salience estimate that complements visual-similarity-based redundancy. The bank is instantiated once at model load, shared across all videos and queries, and each $s(v)$ is computed once per token at encoding time and reused across every subsequent tier transition, so the prior adds only one MaxSim per frame to the streaming cost. Detailed pseudo questions are in Appendix~\ref{sup:pseudo}.

%% algorithm starts
\begin{algorithm}[t]
\caption{\methodname: two-stage streaming visual memory.}
\label{alg:methodname}
\begin{algorithmic}[1]
\Require streaming frames $\{f_t\}$, user query $q$ at $t{=}T$, pseudo-question $Q$, tier capacities $S, M$, token budget $B$, retrieval budget $K$
\Statex \textbf{Stage~1 --- Streaming Memory Generation} \hfill \emph{(query-agnostic, online for $t \le T$)}
\State initialize $\mathcal{M}_{\mathrm{short}}, \mathcal{M}_{\mathrm{mid}}, \mathcal{M}_{\mathrm{long}} \gets \varnothing$
\For{each incoming frame $f_t$}
    \State encode $V_t$, compute $\mathbf{s}_t \gets \operatorname{MaxSim}(V_t, Q)$, and append $(V_t, \mathbf{s}_t)$ to $\mathcal{M}_{\mathrm{short}}$
    \State \textbf{if} $|\mathcal{M}_{\mathrm{short}}| > S$ \textbf{then} evict oldest to $\mathcal{M}_{\mathrm{mid}}$ via \textsc{Temporal Semantic Prune}
    \State \textbf{if} $|\mathcal{M}_{\mathrm{mid}}| > M$ \textbf{then} evict oldest to $\mathcal{M}_{\mathrm{long}}$ via \textsc{Spatial Semantic Select}
    \While{$\mathrm{NumTokens}(\mathcal{M}_{\mathrm{short}} \cup \mathcal{M}_{\mathrm{mid}} \cup \mathcal{M}_{\mathrm{long}}) > B$}
        \State discard the lowest-$\mathbf{s}$ token in $\mathcal{M}_{\mathrm{long}}$
    \EndWhile
\EndFor
\Statex \textbf{Stage~2 --- Adaptive Memory Retrieval} \hfill \emph{(query-aware, at $t{=}T$)}
\State \textbf{if} $\operatorname{MaxSim}(\mathcal{M}_{\mathrm{short}}, q) \ge \rho \cdot \overline{\sigma}^{\,\mathrm{ema}}_{\mathrm{short}}$ \textbf{then} \Return $\mathcal{M}_{\mathrm{short}}$ \Comment{recency gate}
\State compute $\sigma(g) \gets \operatorname{MaxSim}(g, q)$ pooled over tokens, $\forall g \in \mathcal{M}_{\mathrm{mid}} \cup \mathcal{M}_{\mathrm{long}}$
\State $\mathcal{R} \gets$ top-$K$ frames of $\mathcal{M}_{\mathrm{mid}} \cup \mathcal{M}_{\mathrm{long}}$ ranked by $\sigma$; \Return $\mathcal{M}_{\mathrm{short}} \cup \mathcal{R}$
\end{algorithmic}
\end{algorithm}
%% algorithm ends

\paragraph{Three-tier Memory with Selective Forgetting.}
Newly encoded tokens enter $\mathcal{M}_{\mathrm{short}}$ as a first-in-first-out FIFO buffer and are preserved in full to anchor the most recent context. When the window fills, the oldest frame is evicted to $\mathcal{M}_{\mathrm{mid}}$ via Temporal Semantic Pruning, which removes tokens well-represented by their temporal neighbours~\cite{xie2026fluxmem} while sparing those with high $s(v)$ or located at scene boundaries. The semantic prior thus serves as an additional retention criterion that augments the similarity-based baseline without overriding it. When $\mathcal{M}_{\mathrm{mid}}$ overflows, frames pass to $\mathcal{M}_{\mathrm{long}}$ via Spatial Semantic Selection, which treats spatial coverage as a hard constraint and uses $s(v)$ as the ranking criterion within it, so retained tokens remain spatially dispersed yet semantically ordered. All retained tokens carry their original hidden states throughout the hierarchy, and no synthetic or averaged representation is introduced. Beyond per-tier capacities, a Selective Forgetting step enforces the global budget $B$. Once exceeded, the lowest-scoring tokens in $\mathcal{M}_{\mathrm{long}}$ are evicted until the budget is restored, yielding an $\mathcal{O}(1)$ memory footprint independent of video length.

%% ================================================================
\subsection{Stage~2: Adaptive Memory Retrieval}
\label{sec:phase2}
Upon query arrival at $t{=}T$, Stage~2 takes the frozen Stage~1 memory together with the user query $q$ and returns a query-conditioned subset of frames for the downstream MLLM. The module is training-free and inherits the MaxSim primitive from Stage~1 for query-memory scoring. Two design choices, an asymmetric treatment of the recency anchor and an adaptive retrieval scope, enable Stage~2 to accommodate queries that vary in temporal target and evidence distribution.

\paragraph{Anchor-conditioned Recency Gate.}
Streaming queries differ in their temporal target. Some refer to the most recent context, while others probe events distributed over the long history. Motivated by the recency prior reported in prior streaming work~\cite{shen2026simple}, we first determine whether the aged tiers need to be consulted at all. Specifically, we measure the affinity between $q$ and the recency anchor $\mathcal{M}_{\mathrm{short}}$, and compare it against a running statistic of past affinities accumulated during streaming. If the current affinity falls within this range, $\mathcal{M}_{\mathrm{short}}$ is returned directly and retrieval over the aged tiers is bypassed. Calibrating against the anchor's own history removes the need for a query-independent threshold and induces a per-query gate that adapts to the affinity distribution of each video.

\paragraph{Query-aware Late Interaction.}
When the gate routes the query to the aged tiers, the query is encoded with the same tokenizer used for the pseudo-question bank in Stage~1, so query tokens and visual hidden states already lie in a shared representation space and can be compared directly without an additional projection or encoding pass. For each candidate frame $g \in \mathcal{M}_{\mathrm{mid}} \cup \mathcal{M}_{\mathrm{long}}$, we compute a ColBERT-style late-interaction~\cite{khattab2020colbert} score
\begin{equation}
\sigma(g) \;=\; \frac{1}{|g|} \sum_{i=1}^{|g|}
\max_{q_j \in q}\; \cos\bigl(g_i,\, q_j\bigr),
\label{eq:sigma}
\end{equation}
where each retained visual token contributes its maximum cosine similarity against the query and the frame relevance is the mean of these per-token maxima. The decision is thus made at frame level while the evidence is aggregated from token level, which matches the granularity at which Stage~1 operates on visual redundancy and at which queries typically refer to frame-spanning events. Reusing the MaxSim primitive across both stages also means that query-agnostic compression and query-aware retrieval share a single scoring mechanism.

\paragraph{Adaptive Retrieval with Anchor Preservation.}
We rank $\mathcal{M}_{\mathrm{mid}} \cup \mathcal{M}_{\mathrm{long}}$ by $\sigma(\cdot)$ and retain a query-conditioned subset $\mathcal{R}$ whose size adapts to the dispersion of the relevance scores. Well-separated scores indicate that a small number of frames are clearly more relevant and trigger aggressive filtering, while tightly clustered scores indicate diffuse evidence and cause retrieval to keep more candidates, which is useful for counting or coverage-style queries where signal is spread across many frames. The recency anchor $\mathcal{M}_{\mathrm{short}}$ is appended unconditionally, yielding the final retrieved set $\mathcal{M}^{\star} = \mathcal{M}_{\mathrm{short}} \cup \mathcal{R}$ that is forwarded to the MLLM. This asymmetry reflects the design split between the two stages. The anchor holds recent input at full fidelity and covers present-oriented queries, whereas the aged tiers span a long history where most frames are irrelevant to any given query and benefit from adaptive filtering conditioned on $q$.

\section{Experiments}

\begin{table*}[t]
  \centering
  \setlength{\tabcolsep}{4pt}
  \renewcommand{\arraystretch}{1.2}
  \resizebox{\textwidth}{!}{
  \begin{tabular}{l c c @{\hspace{6pt}{\color{gray!60}\vrule width 0.3pt}\hspace{6pt}} c c c c c c @{\hspace{6pt}{\color{gray!60}\vrule width 0.3pt}\hspace{6pt}} c @{\hspace{6pt}{\color{gray!60}\vrule width 0.3pt}\hspace{6pt}} c c c @{\hspace{6pt}{\color{gray!60}\vrule width 0.3pt}\hspace{6pt}} c @{\hspace{6pt}{\color{gray!60}\vrule width 0.3pt}\hspace{6pt}} c}
    \toprule
    \multirow{2}{*}{Method} & \multirow{2}{*}{Size} & \multirow{2}{*}{Frames} & \multicolumn{7}{c}{Real-Time Visual Perception} & \multicolumn{4}{c}{Backward Tracing} & \multirow{2}{*}{Overall} \\
    \cmidrule(lr){4-10} \cmidrule(lr){11-14}
    &&& OCR & ACR & ATR & STU & FPD & OJR & Avg. & EPM & ASI & HLD & Avg. & \\
    \midrule

    \rowcolor{isabelline}
    \multicolumn{15}{c}{\textit{Proprietary Models}} \\
    \midrule
    Gemini 1.5 Pro~\cite{team2024gemini} & -- & 1~fps & 85.91 & 66.97 & 79.31 & 58.43 & 63.37 & 61.96 & 69.32 & 58.59 & 76.35 & 52.64 & 62.54 & 65.93 \\
    GPT-4o~\cite{hurst2024gpt4o} & -- & 64 & 69.80 & 64.22 & 71.55 & 51.12 & 70.30 & 59.78 & 64.46 & 57.91 & 75.68 & 48.66 & 60.75 & 62.60 \\
    \midrule

    \rowcolor{isabelline}
    \multicolumn{15}{c}{\textit{Open-source Offline MLLMs}} \\
    \midrule
    LLaVA-Video~\cite{zhang2024llavavideo} & 7B & 64 & 69.80 & 59.63 & 66.38 & 50.56 & \underline{72.28} & 61.41 & 63.34 & 51.18 & 64.19 & 9.68 & 41.68 & 52.51 \\
    Qwen2-VL~\cite{wang2024qwen2} & 7B & 64 & 69.13 & 53.21 & 63.79 & 50.56 & 66.34 & 60.87 & 60.65 & 44.44 & \textbf{66.89} & 34.41 & 48.58 & 54.61 \\
    InternVL2~\cite{chen2024internvl2} & 8B & 64 & 68.46 & 58.72 & 68.97 & 44.94 & 67.33 & 55.98 & 60.73 & 43.10 & 61.49 & 27.41 & 44.00 & 52.36 \\
    LongVU~\cite{shen2024longvu} & 7B & 1~fps & 55.70 & 49.54 & 59.48 & 48.31 & 68.32 & 63.04 & 57.40 & 43.10 & \underline{66.22} & 9.14 & 39.49 & 48.45 \\
    \midrule

    \rowcolor{isabelline}
    \multicolumn{15}{c}{\textit{Open-source Online MLLMs (Training-Based)}} \\
    \midrule
    VideoLLM-Online~\cite{chen2024videollm}~\textsubscript{\textcolor{grey}{[CVPR 2024]}} & 8B & 2~fps & 8.05 & 23.85 & 12.07 & 14.04 & 45.54 & 21.20 & 20.79 & 22.22 & 18.80 & 12.18 & 17.73 & 19.26 \\
    Dispider~\cite{qian2025dispider}~\textsubscript{\textcolor{grey}{[CVPR 2025]}} & 7B & 1~fps & 57.72 & 49.54 & 62.07 & 44.94 & 61.39 & 51.63 & 54.55 & 48.48 & 55.41 & 4.30 & 36.06 & 45.30 \\
    Flash-VStream~\cite{zhang2025flash}~\textsubscript{\textcolor{grey}{[ICCV 2025]}} & 7B & 1~fps & 25.50 & 32.11 & 29.31 & 33.71 & 29.70 & 28.80 & 29.86 & 36.36 & 33.78 & 5.91 & 25.35 & 27.61 \\
    ViSpeak~\cite{fu2025vispeak}~\textsubscript{\textcolor{grey}{[ICCV 2025]}} & 7B & 1~fps & 75.20 & 58.72 & 71.55 & 51.12 & \textbf{74.26} & 66.85 & 66.30 & \textbf{59.93} & 48.65 & \textbf{63.98} & \textbf{57.52} & \underline{61.91} \\
    ThinkStream~\cite{liu2026thinking} & 3B & 1~fps & 85.23 & 64.22 & 69.83 & 49.44 & 69.31 & 64.13 & 67.03 & 53.87 & 59.46 & \underline{43.55} & \underline{52.30} & 59.66 \\
    TimeChat-Online~\cite{yao2025timechat}~\textsubscript{\textcolor{grey}{[ACM MM 2025]}} & 7B & 1~fps & 75.20 & 46.80 & 70.70 & 47.80 & 69.30 & 61.40 & 61.90 & 55.90 & 59.50 & 9.70 & 41.70 & 51.80 \\
    StreamForest~\cite{zeng2025streamforest}~\textsubscript{\textcolor{grey}{[NeurIPS 2025]}} & 7B & 1~fps & 68.46 & 53.21 & 71.55 & 47.75 & 65.35 & 60.87 & 61.20 & \underline{58.92} & 64.86 & 32.26 & 52.02 & 56.61 \\
    \midrule

    \rowcolor{isabelline}
    \multicolumn{15}{c}{\textit{Open-source Online MLLMs (Training-Free)}} \\
    \midrule

    Qwen2.5-VL-3B$^{\dagger}$~\cite{bai2025qwen2_5vl} & 3B & 1~fps & 77.18 & 52.29 & 68.97 & 41.01 & 67.33 & 60.87 & 61.27 & 49.83 & 53.38 & 26.34 & 43.18 & 52.23 \\
    \rowcolor{lightblue}
    \hspace{3pt} \textbf{+ \methodname\ (Ours)} & 3B & 1~fps & \underline{89.26} & \underline{68.81} & \underline{73.28} & \underline{55.62} & 64.36 & \underline{69.57} & \underline{70.15} & 47.81 & 56.76 & 29.03 & 44.53 & 57.34~\small\textcolor{teal}{(+5.12)} \\
    \midrule

    Qwen2.5-VL-7B$^{\dagger}$~\cite{bai2025qwen2_5vl} & 7B & 1~fps & 67.79 & 55.05 & 67.24 & 42.13 & 66.34 & 60.87 & 59.90 & 51.52 & 58.78 & 23.66 & 44.65 & 52.27 \\
    \rowcolor{lightblue}
    \hspace{3pt} \textbf{+ \methodname\ (Ours)} & 7B & 1~fps & \textbf{91.95} & \textbf{68.81} & \textbf{81.03} & \textbf{65.73} & 70.30 & \textbf{71.74} & \textbf{74.93} & 50.84 & 62.84 & 37.63 & 50.44 & \textbf{62.69}~\small\textcolor{teal}{(+10.41)} \\

    \bottomrule
  \end{tabular}
  }
    \caption{Comparison with state-of-the-art Methods on OVO-Bench. \textbf{Bold} indicates the best and \underline{underline} indicates the second best among all open-source models. \small $^{\dagger}$ indicates the reproduced results.}
    \label{tab_ovo_free}
\end{table*}

\subsection{Experimental Setup}

\paragraph{Benchmark}
We evaluate our method on three benchmarks for online streaming video understanding. From \textbf{OVO-Bench}~\cite{niu2025ovo}, which emphasizes timestamp-aware reasoning along the video timeline, we use the Real-Time Visual Perception and Backward Tracing subsets as our primary evaluation. The Real-Time Visual Perception subset contains 837 QA pairs across 6 tasks, assessing a model's ability to perceive and respond to events at the current timestamp, covering optical character recognition (OCR), action recognition (ACR), attribute recognition (ATR), spatial understanding (STU), future prediction (FPD), and object recognition (OJR). The Backward Tracing subset contains 631 QA pairs across 3 tasks, requiring the model to trace back to past events to answer the current query, covering episodic memory (EPM), action sequence identification (ASI), and hallucination detection (HLD). From \textbf{StreamingBench}~\cite{lin2024streamingbench}, we adopt the Real-Time Visual Understanding subset with 2,500 QA pairs, which evaluates the perception of visual changes in streaming inputs. Finally, we use the Static and Dynamic subsets of \textbf{ODV-Bench}~\cite{zeng2025streamforest}, a benchmark for online streaming video understanding in autonomous driving, covering fine-grained object and action recognition, spatial relation description, trajectory prediction, and risk event assessment.

\paragraph{Implementation Details}
We apply \methodname\ to the frozen Qwen2.5-VL-7B-Instruct backbone in a
training-free manner. Videos are sampled at 1\,FPS up to 256 frames with
a $512{\times}28{\times}28$ visual-token budget per frame. Stage~1
maintains a 4-frame recency anchor and a 16-frame mid-term tier under a
global token budget of $B{=}2048$. Stage~2 returns the top-$K$
non-anchor frames ranked by Eq.~(\ref{eq:sigma}); the recency-gate
ratio $\rho$ is the only task-dependent hyperparameter, set to
$\rho{=}0.1$ for real-time queries and $\rho{=}2.0$ for backward queries
that require historical retrieval. All experiments run on a single
NVIDIA GPU.

\begin{table*}[t]
  \centering
  \setlength{\tabcolsep}{4pt}
  \renewcommand{\arraystretch}{1.2}
  \resizebox{\textwidth}{!}{
  \begin{tabular}{l c @{\hspace{6pt}{\color{gray!60}\vrule width 0.3pt}\hspace{6pt}} c c c c c c @{\hspace{6pt}{\color{gray!60}\vrule width 0.3pt}\hspace{6pt}} c @{\hspace{6pt}{\color{gray!60}\vrule width 0.3pt}\hspace{6pt}} c c c @{\hspace{6pt}{\color{gray!60}\vrule width 0.3pt}\hspace{6pt}} c @{\hspace{6pt}{\color{gray!60}\vrule width 0.3pt}\hspace{6pt}} c @{\hspace{6pt}{\color{gray!60}\vrule width 0.3pt}\hspace{6pt}}}
    \toprule
    \multirow{2}{*}{Method} & \multirow{2}{*}{Frames} & \multicolumn{7}{c}{Real-Time Visual Perception} & \multicolumn{4}{c}{Backward Tracing} & \multirow{2}{*}{Overall} \\
    \cmidrule(lr){3-9} \cmidrule(lr){10-13}
    && OCR & ACR & ATR & STU & FPD & OJR & Avg. & EPM & ASI & HLD & Avg. & \\
    \midrule
    \rowcolor{isabelline}
    \multicolumn{14}{c}{\textit{Backbone: Qwen2.5-VL-3B}} \\
    \midrule
    Qwen2.5-VL-3B$^{\dagger}$~\cite{bai2025qwen2_5vl} & 1~fps & 77.18 & 52.29 & \underline{68.97} & 41.01 & \underline{67.33} & 60.87 & 61.27 & \textbf{49.83} & 53.38 & \underline{26.34} & \underline{43.18} & 52.23 \\
    \hspace{3pt} + FluxMem~\cite{xie2026fluxmem}~\textsubscript{\textcolor{grey}{[CVPR 2026]}} & 1~fps & \underline{83.22} & \underline{56.88} & 67.24 & \underline{47.75} & \textbf{68.32} & \underline{63.59} & \underline{64.50} & 47.47 & \underline{54.73} & 24.19 & 42.13 & \underline{53.31} \\
    \rowcolor{lightblue}
    \hspace{3pt} \textbf{+ \methodname\ (Ours)} & 1~fps & \textbf{89.26} & \textbf{68.81} & \textbf{73.28} & \textbf{55.62} & 64.36 & \textbf{69.57} & \textbf{70.15} & \underline{47.81} & \textbf{56.76} & \textbf{29.03} & \textbf{44.53} & \textbf{57.34} \\
    \midrule
    \rowcolor{isabelline}
    \multicolumn{14}{c}{\textit{Backbone: Qwen2.5-VL-7B}} \\
    \midrule
    Qwen2.5-VL-7B$^{\dagger}$~\cite{bai2025qwen2_5vl} & 1~fps & 67.79 & 55.05 & 67.24 & 42.13 & 66.34 & 60.87 & 59.90 & \textbf{51.52} & 58.78 & 23.66 & 44.65 & 52.27 \\
    \hspace{3pt} + HERMES (6K)~\cite{zhang2026hermes} & 0.5~fps & \underline{85.91} & 60.55 & \underline{74.14} & 52.81 & 70.30 & \underline{66.85} & 68.42 & 49.49 & 58.78 & \underline{33.33} & 48.10 & 58.26 \\
    \hspace{3pt} + HERMES (4K)~\cite{zhang2026hermes} & 0.5~fps & 85.23 & \underline{64.22} & 71.55 & \underline{53.37} & \underline{74.26} & 65.22 & \underline{68.98} & 48.48 & \underline{62.16} & \textbf{37.63} & \underline{49.43} & \underline{59.20} \\
    \hspace{3pt} + FluxMem~\cite{xie2026fluxmem}~\textsubscript{\textcolor{grey}{[CVPR 2026]}} & 1~fps & 81.21 & 59.63 & 70.69 & 53.37 & \textbf{75.25} & 63.04 & 67.20 & 48.48 & \textbf{64.19} & 29.03 & 47.24 & 57.22 \\
    \rowcolor{lightblue}
    \hspace{3pt} \textbf{+ \methodname\ (Ours)} & 1~fps & \textbf{91.95} & \textbf{68.81} & \textbf{81.03} & \textbf{65.73} & 70.30 & \textbf{71.74} & \textbf{74.93} & \underline{50.84} & 62.84 & \textbf{37.63} & \textbf{50.44} & \textbf{62.69} \\
    \bottomrule
  \end{tabular}
  }
    \caption{Comparison with Other Training-free Methods on OVO-Bench. \textbf{Bold} indicates the best and \underline{underline} indicates the second best within each backbone group. \small $^{\dagger}$ indicates the reproduced results.}
    \label{tab_ovobench}
\end{table*}

\paragraph{OVO-Bench.}

Tables~\ref{tab_ovo_free} and~\ref{tab_ovobench} report results on OVO-Bench. \methodname\ attains the highest overall scores on both backbones, 62.69 on Qwen2.5-VL-7B and 57.34 on Qwen2.5-VL-3B, comparing favorably with training-free competitors and with training-based approaches such as ViSpeak at 61.91 and StreamForest at 56.61. The improvements are most pronounced on real-time visual perception, where the average rises by 15.03 points on 7B and 8.88 points on 3B, with consistent gains across OCR, STU, ACR, ATR, and OJR. These subtasks rely on fine-grained visual details concentrated in a small number of frames, which aligns with the design of the pseudo-question semantic prior in Stage~1, intended to retain such tokens during compression rather than treating them as visually redundant. On backward tracing, \methodname\ also yields the best average on both backbones, 50.44 on 7B and 44.53 on 3B, with hallucination detection improving by 13.97 on 7B and 2.69 on 3B. This is consistent with the role of Stage~2, which forwards a query-conditioned subset of frames to the MLLM and thereby limits exposure to non-relevant visual evidence. Episodic memory shows a small decrease on both backbones, from 49.83 to 47.81 on 3B and from 51.52 to 50.84 on 7B, reflecting a trade-off inherent to bounded-memory retrieval, where target episodes that are visually dissimilar to the query may receive low relevance scores under late interaction and be filtered by adaptive top-K selection. The case study in Fig.~\ref{fig:teaser} illustrates this behavior on a query asking what the person did before preparing meat. Stage~1 retains seasoning-preparation frames in the mid-term tier while pushing visually redundant frames into the long-term tier. Recency gate in Stage~2 then routes the query past the short-term anchor that holds the meat-preparation content, allowing late interaction over the mid-term and long-term tier to surface the seasoning step that directly supports the correct answer.

\begin{figure}
    \centering
    \includegraphics[width=1.0\linewidth]{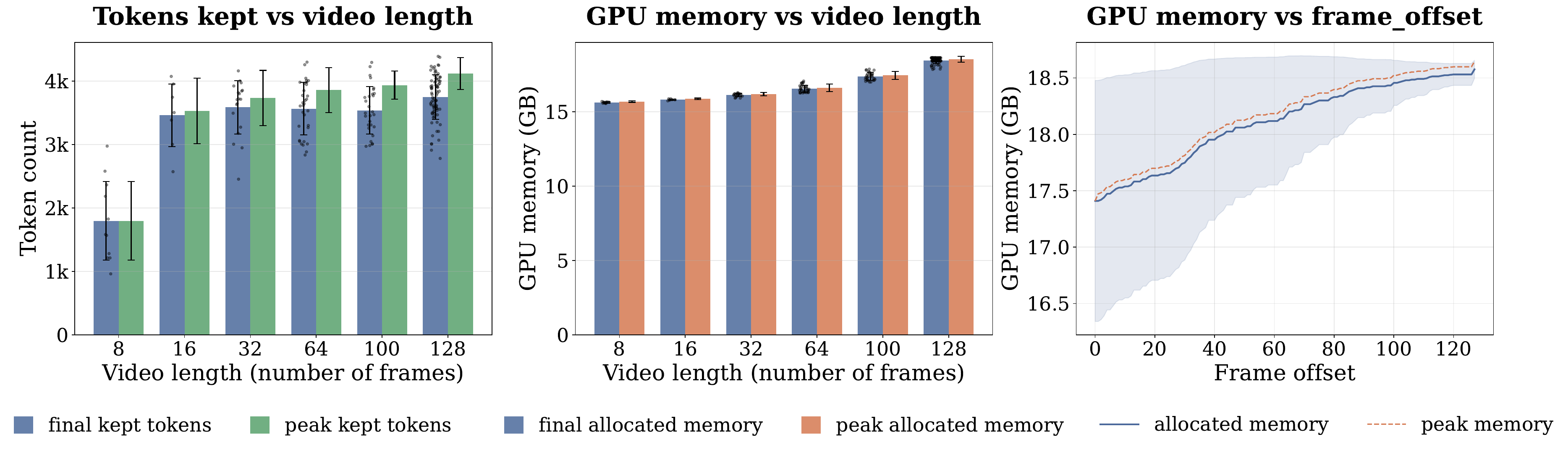}
    \caption{\textbf{Efficiency Analysis} of \methodname. The figure shows retained token counts and GPU memory across video lengths from 8 to 128 frames, as well as the memory trajectory along frame offsets within a single 128-frame run.}
    \label{fig:efficiency}
\end{figure}

\paragraph{StreamingBench and ODV-Bench.}
On StreamingBench and ODV-Bench, \methodname\ delivers consistent improvements across both backbones. On the 7B backbone, the StreamingBench Real-time score rises from 73.9 to 76.0, the ODV-Bench Static score from 48.3 to 57.0, and the ODV-Bench Dynamic score from 57.5 to 60.7. On the 3B backbone, the corresponding improvements are 1.8, 1.9, and 0.7 points. The largest gain, 8.7 points on the 7B Static category, falls on a setting that depends on preserving fine-grained visual details under compression, which is consistent with the design of the semantic prior and Spatial Semantic Selection in Stage~1.

\subsection{Analysis}

\paragraph{Efficiency Analysis.} Fig.~\ref{fig:efficiency} reports the token and memory footprint of \methodname\ as video length scales from 8 to 128 frames, providing direct empirical evidence for the design of Stage 1. As the input length grows by 16$\times$
, the number of retained tokens increases only from roughly 1.8k to 3.7k, a sub-linear trend that reflects the joint effect of Temporal Semantic Pruning, Spatial Semantic Selection, and the global budget enforced by Selective Forgetting. Correspondingly, peak GPU memory rises mildly from 15.5\,GB to 18.5\,GB, in contrast to the 35.8\,GB consumed by the Qwen2.5-VL-7B baseline at 128 frames, amounting to a 48\% reduction at the longest setting. The peak and final allocations remain nearly indistinguishable across all lengths, indicating that incoming frames are compressed into the existing budget on the fly rather than buffered at full resolution before pruning. Within a single 128-frame run, the memory trajectory follows a concave, fast-saturating curve with a narrowing variance band, suggesting that the three-tier cascade quickly converges to a steady state regardless of input content. Taken together, these results show that Stage 1 transforms long-video processing from a length-bounded into a budget-bounded problem, which is what makes the query-aware retrieval in Stage 2 tractable on commodity GPUs.

\begin{table}[t]
  \centering
  \setlength{\tabcolsep}{3pt}
  \renewcommand{\arraystretch}{1.15}
  \small
  \begin{tabular}{l @{\hspace{6pt}{\color{gray!60}\vrule width 0.3pt}\hspace{6pt}} c @{\hspace{6pt}{\color{gray!60}\vrule width 0.3pt}\hspace{6pt}} c c}
    \toprule
    \multirow{2}{*}{Method} & \textbf{StreamingBench} & \multicolumn{2}{c}{\textbf{ODV-Bench}} \\
    \cmidrule(lr){2-2} \cmidrule(lr){3-4}
    & Real-time & Static & Dynamic \\
    \midrule
    LongVA~\cite{zhang2024longva}          & 60.0 & 31.8 & 56.6 \\
    LLaVA-OV~\cite{li2024llava_ov} & 71.1 & 34.2 & 55.1 \\
    Flash-VStream~\cite{zhang2025flash}   & 23.2 & 24.8 & 40.2 \\
    Dispider~\cite{qian2025dispider}        & 67.6 & 32.5 & 52.7 \\
    \midrule
    Qwen2.5-VL-3B & 68.9 & 46.0 & 55.9 \\
    \rowcolor{lightblue}
    \textbf{+ \methodname\ (Ours)} & 70.7\,\textcolor{teal}{\footnotesize(+1.8)} & 47.9\,\textcolor{teal}{\footnotesize(+1.9)} & 56.6\,\textcolor{teal}{\footnotesize(+0.7)} \\
    \midrule
    Qwen2.5-VL-7B & 73.9 & 48.3 & 57.5 \\
    \rowcolor{lightblue}
    \textbf{+ \methodname\ (Ours)} & 76.0\,\textcolor{teal}{\footnotesize(+2.1)} & 57.0\,\textcolor{teal}{\footnotesize(+8.7)} & 60.7\,\textcolor{teal}{\footnotesize(+3.2)} \\
    \bottomrule
  \end{tabular}
  \caption{Evaluation on StreamingBench and ODV-Bench.}
  \label{tab:results_combined}
\end{table}

\begin{figure}  % r 表示右边,也可用 l
    \centering
    \includegraphics[width=0.7\textwidth]{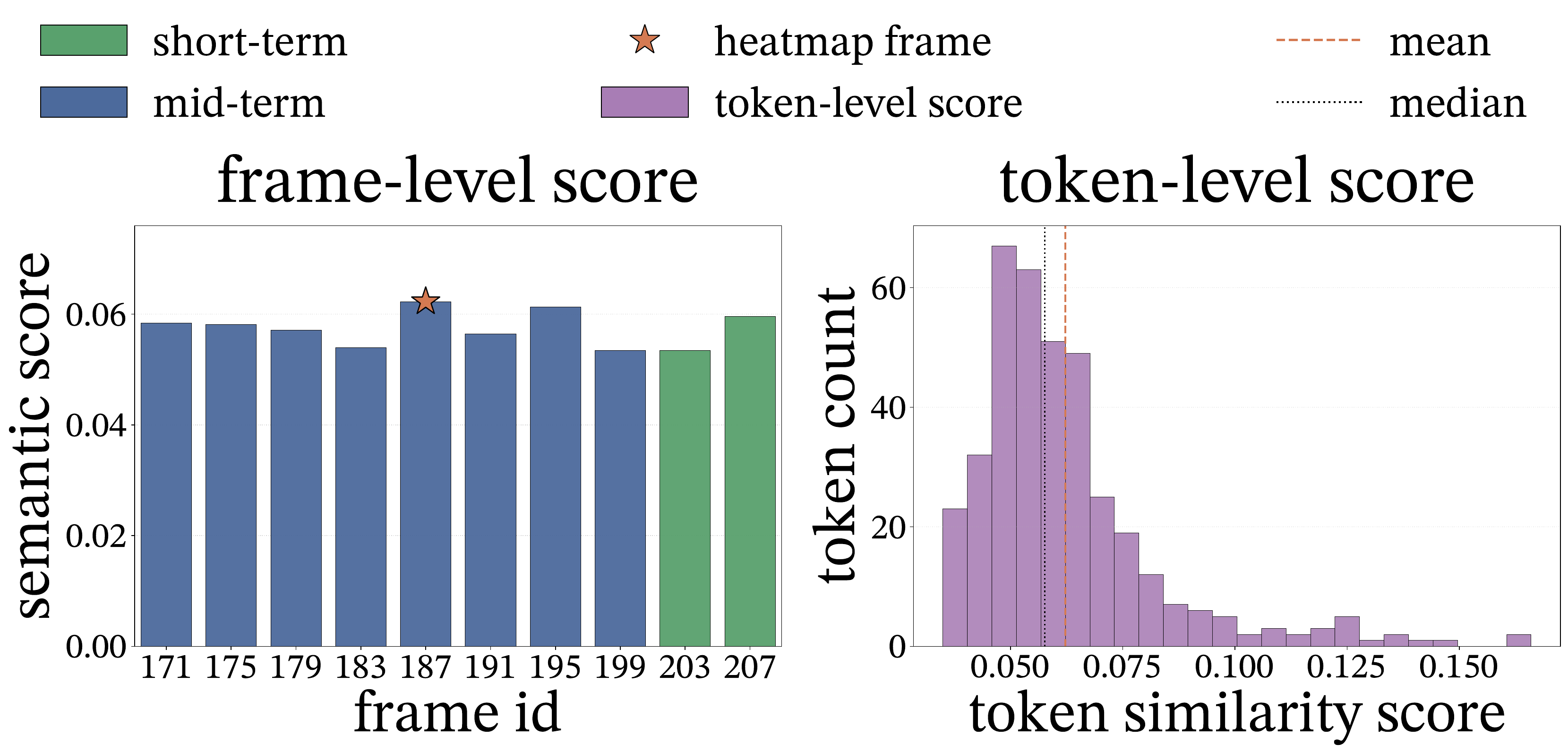}
    \caption{\textbf{Semantic score distributions} from the pseudo-question bank at the frame level (left) and token level within a single frame (right).}
    \label{fig:score}
\end{figure}

\paragraph{Score Analysis.} Fig.~\ref{fig:score} visualizes the semantic score distribution produced by the pseudo-question bank during a representative inference step, offering a closer view of how the query-agnostic prior behaves in practice. At the frame level, scores across short-term and mid-term frames are tightly clustered within roughly 0.054 to 0.063, with no systematic gap between the two tiers. This narrow spread indicates that the prior provides only weak frame-wise discrimination, consistent with our choice of performing Temporal Semantic Pruning by relative ranking within a sliding window rather than by absolute thresholds, since neighboring frames in a continuous video are inherently similar in semantics. The token-level distribution within a single frame tells a different story: it is heavily right-skewed, with most tokens concentrated near 0.05 and a small high-score tail extending beyond 0.15 that pulls the mean above the median. This long-tailed structure aligns with the rationale behind Spatial Semantic Selection, where retaining a small subset of high-scoring tokens preserves the informative content of a frame while discarding the low-score majority reduces intra-frame redundancy. The contrast between the two distributions further illustrates that temporal and spatial informativeness exhibit fundamentally different statistical structures, which motivates applying distinct selection operators at the two granularities within our cascade.

\subsection{Ablation Study}

We conduct three groups of ablations on OVO-Bench with Qwen2.5-VL-7B, where Avg. denotes the unweighted mean of the Real-Time Visual Perception and Backward Tracing subsets. Table~\ref{tab:ablation_components}(a) validates the two-stage decomposition. Stage 1 alone lifts Avg. from 52.27 to 57.79 by compressing redundant visual tokens through hierarchical forgetting, while Stage 2 alone reaches 59.95 via query-aware retrieval. Their combination further attains 62.69, indicating that query-agnostic memory construction and query-aware retrieval contribute complementary benefits. Table~\ref{tab:ablation_components}(b) verifies the pseudo-question prior. Random vectors yield only 56.05 as semantically unstructured priors degenerate into near-random dropping, and a single fixed prompt reaches 59.73 but suffers from single-view coverage bias. Our prompt bank instead achieves 62.69 by aggregating multi-view priors that better approximate the latent query distribution. Table~\ref{tab:ablation_components}(c) examines the recency gate. The No gate variant attains 59.06, while the Always gate variant peaks at 73.62 on rt. but collapses to 45.18 on bw., since over-emphasizing recency suppresses the long-range evidence required by backward tracing. Our EMA gate instead adaptively balances recency and long-range fidelity, reaching 74.93 on rt., 50.44 on bw., and 62.69 on Avg. Taken together, these results indicate that staged decomposition, multi-view priors, and adaptive gating each contribute to the performance gains of our framework.

\begin{table*}[t]
  \centering
  \footnotesize
  \setlength{\tabcolsep}{3pt}
  \renewcommand{\arraystretch}{1.1}
  \begin{minipage}[t]{0.30\linewidth}
    \centering
    \textit{(a) Two-stage decomposition} \\[2pt]
    \begin{tabular}{l c c c}
      \toprule
      Variant & rt. & bw. & Avg. \\
      \midrule
      baseline           & 59.90 & 44.65 & 52.27 \\
      \midrule
      Stage 1 only       & 68.16 & 47.42 & 57.79 \\
      Stage 2 only       & 71.07 & 48.83 & 59.95 \\
      \rowcolor{lightblue}
      S1 + S2 (ours)     & \textbf{74.93} & \textbf{50.44} & \textbf{62.69} \\
      \bottomrule
    \end{tabular}
  \end{minipage}
  \hspace{0.04\linewidth}
  \begin{minipage}[t]{0.30\linewidth}
    \centering
    \textit{(b) Pseudo-question prior $Q$} \\[2pt]
    \begin{tabular}{l c c c}
      \toprule
      Variant & rt. & bw. & Avg. \\
      \midrule
      baseline           & 59.90 & 44.65 & 52.27 \\
      \midrule
      Random vectors     & 65.78 & 46.32 & 56.05 \\
      Single prompt      & 70.51 & 48.95 & 59.73 \\
      \rowcolor{lightblue}
      Bank (ours)        & \textbf{74.93} & \textbf{50.44} & \textbf{62.69} \\
      \bottomrule
    \end{tabular}
  \end{minipage}
  \hspace{0.04\linewidth}
  \begin{minipage}[t]{0.30\linewidth}
    \centering
    \textit{(c) Recency gate} \\[2pt]
    \begin{tabular}{l c c c}
      \toprule
      Variant & rt. & bw. & Avg. \\
      \midrule
      baseline           & 59.90 & 44.65 & 52.27 \\
      \midrule
      No gate            & 69.84 & 48.27 & 59.06 \\
      Always gate        & 73.62 & 45.18 & 59.40 \\
      \rowcolor{lightblue}
      EMA (ours)         & \textbf{74.93} & \textbf{50.44} & \textbf{62.69} \\
      \bottomrule
    \end{tabular}
  \end{minipage}
    \caption{Component ablations on OVO-Bench with Qwen2.5-VL-7B. ``rt.'' / ``bw.'' denote the Real-Time Visual Perception and Backward Tracing subset averages; ``Avg.'' is the unweighted mean of rt. and bw. The first row in each panel is the raw backbone for reference. Default settings are highlighted; best results within each ablation group are in \textbf{bold}.}
    \label{tab:ablation_components}
\end{table*}

\section{Limitation}
While \methodname\ delivers consistent improvements across benchmarks, several limitations remain. First, under a bounded memory budget, queries requiring exhaustive temporal coverage such as counting and causal reasoning are inherently sensitive to the trade-off between selective retrieval and full coverage, which is a property shared by bounded-memory streaming methods in general. Second, the pseudo-question bank consists of a fixed set of generic probes, and adapting it to specific video domains is left as future work. Third, \methodname\ is training-free by design, which precludes joint optimization across the two stages, and lightweight tuning of the shared scoring module is a natural extension. We view these as natural directions consistent with the overall design of \methodname\ rather than fundamental obstacles.

\section{Conclusion}
We presented \methodname, a training-free dual-stage framework for online streaming video understanding that brings semantic awareness into compression and adapts the retrieval scope per query. Stage 1 builds a three-tier streaming memory online under a constant budget, where a fixed pseudo-question bank shapes long-term retention by semantic salience rather than visual similarity alone. Stage 2 performs query-aware retrieval over this memory via an anchor-conditioned recency gate and late interaction, with the two stages sharing a single MaxSim primitive so that compression and retrieval are coordinated as one pipeline. Applied to Qwen2.5-VL without any training, \methodname\ improves the OVO-Bench overall score from 52.27 to 62.69, yields consistent gains on StreamingBench and ODV-Bench, and reduces peak GPU memory by 48\% at 128 frames. These results suggest that decoupling query-agnostic semantic compression from query-aware adaptive retrieval is a practical direction for scaling streaming video understanding without training.

\section*{Acknowledgement}
The work is partially supported by the U.S. National Science Foundation (NSF) Grant CRII 2451683, a U.S. Bank Academic Research Award, the University of California, Merced, and a UC Merced Faculty Research Award.

\clearpage

\bibliographystyle{plainnat}
\bibliography{main}

@String(CVPR= {IEEE Conf. Comput. Vis. Pattern Recog.})

@String(ICCV= {Int. Conf. Comput. Vis.})

@String(ICLR = {Int. Conf. Learn. Represent.})

@String(CVPR  = {CVPR})

@String(ICCV  = {ICCV})

@String(ICLR  = {ICLR})

@article{zhang2024llavavideo,
  title={LLaVA-Video: Video Instruction Tuning With Synthetic Data},
  author={Zhang, Yuanhan and Wu, Jinming and Li, Wei and Li, Bo and Ma, Zejun and Liu, Ziwei and Li, Chunyuan},
  journal={arXiv preprint arXiv:2410.02713},
  year={2024}
}

@article{li2024llava_ov,
  title={Llava-onevision: Easy visual task transfer},
  author={Li, Bo and Zhang, Yuanhan and Guo, Dong and Zhang, Renrui and Li, Feng and Zhang, Hao and Zhang, Kaichen and Zhang, Peiyuan and Li, Yanwei and Liu, Ziwei and others},
  journal={arXiv preprint arXiv:2408.03326},
  year={2024}
}

@article{wang2023chatvideo,
  title={Chatvideo: A tracklet-centric multimodal and versatile video understanding system},
  author={Wang, Junke and Chen, Dongdong and Luo, Chong and Dai, Xiyang and Yuan, Lu and Wu, Zuxuan and Jiang, Yu-Gang},
  journal={arXiv preprint arXiv:2304.14407},
  year={2023}
}

@article{wang2024qwen2,
  title={Qwen2-vl: Enhancing vision-language model's perception of the world at any resolution},
  author={Wang, Peng and Bai, Shuai and Tan, Sinan and Wang, Shijie and Fan, Zhihao and Bai, Jinze and Chen, Keqin and Liu, Xuejing and Wang, Jialin and Ge, Wenbin and others},
  journal={arXiv preprint arXiv:2409.12191},
  year={2024}
}

@article{bai2025qwen2_5vl,
  title={Qwen2.5-VL Technical Report},
  author={Bai, Shuai and Chen, Keqin and Liu, Xuejing and Wang, Jialin and Ge, Wenbin and Song, Sibo and Dang, Kai and Wang, Peng and Wang, Shijie and Tang, Jun and Zhong, Humen and Zhu, Yuanzhi and Yang, Mingkun and Li, Zhaohai and Wan, Jianqiang and Wang, Pengfei and Ding, Wei and Fu, Zheren and Xu, Yiheng and Ye, Jiabo and Zhang, Xi and Xie, Tianbao and Cheng, Zesen and Zhang, Hang and Yang, Zhibo and Xu, Haiyang and Lin, Junyang},
  journal={arXiv preprint arXiv:2502.13923},
  year={2025}
}

@article{wang2023see,
  title={To see is to believe: Prompting gpt-4v for better visual instruction tuning},
  author={Wang, Junke and Meng, Lingchen and Weng, Zejia and He, Bo and Wu, Zuxuan and Jiang, Yu-Gang},
  journal={arXiv preprint arXiv:2311.07574},
  year={2023}
}

@article{liu2025aligning,
  title={Aligning Cyber Space with Physical World: A Comprehensive Survey on Embodied AI},
  author={Liu, Yang and Chen, Weixing and Bai, Yongjie and Liang, Xiaodan and Li, Guanbin and Gao, Wen and Lin, Liang},
  journal={IEEE/ASME Transactions on Mechatronics},
  year={2025}
}

@article{chen2023e2esurvey,
  title={End-to-end Autonomous Driving: Challenges and Frontiers},
  author={Chen, Li and Wu, Penghao and Chitta, Kashyap and Jaeger, Bernhard and Geiger, Andreas and Li, Hongyang},
  journal={IEEE Transactions on Pattern Analysis and Machine Intelligence},
  year={2024}
}

@article{lee2018interaction,
  title={Interaction methods for smart glasses: A survey},
  author={Lee, Lik-Hang and Hui, Pan},
  journal={IEEE access},
  volume={6},
  pages={28712--28732},
  year={2018},
  publisher={IEEE}
}

@article{ning2025livevlm,
  title={LiveVLM: Efficient Online Video Understanding via Streaming-Oriented KV Cache and Retrieval},
  author={Ning, Zhenyu and Liu, Guangda and Jin, Qihao and Ding, Wenchao and Guo, Minyi and Zhao, Jieru},
  journal={arXiv preprint arXiv:2505.15269},
  year={2025}
}

@article{yang2025streammem,
  title={StreamMem: Query-Agnostic KV Cache Memory for Streaming Video Understanding},
  author={Yang, Yanlai and Zhao, Zhuokai and Shukla, Satya Narayan and Singh, Aashu and Mishra, Shlok Kumar and Zhang, Lizhu and Ren, Mengye},
  journal={arXiv preprint arXiv:2508.15717},
  year={2025}
}

@article{xu2025streamingvlm,
  title={StreamingVLM: Real-Time Understanding for Infinite Video Streams},
  author={Xu, Ruyi and Xiao, Guangxuan and Chen, Yukang and He, Liuning and Peng, Kelly and Lu, Yao and Han, Song},
  journal={arXiv preprint arXiv:2510.09608},
  year={2025}
}

@inproceedings{zhang2025flash,
  title={Flash-vstream: Memory-based real-time understanding for long video streams},
  author={Zhang, Haoji and Wang, Yiqin and Tang, Yansong and Liu, Yong and Feng, Jiashi and Dai, Jifeng and Jin, Xiaojie},
  booktitle={ICCV},
  year={2025}
}

@article{wang2025streambridge,
  title={StreamBridge: Turning Your Offline Video Large Language Model into a Proactive Streaming Assistant},
  author={Wang, Haibo and Feng, Bo and Lai, Zhengfeng and Xu, Mingze and Li, Shiyu and Ge, Weifeng and Dehghan, Afshin and Cao, Meng and Huang, Ping},
  journal={NeurIPS},
  year={2025}
}

@inproceedings{yao2025timechat,
  title={TimeChat-Online: 80\% Visual Tokens are Naturally Redundant in Streaming Videos},
  author={Yao, Linli and Li, Yicheng and Wei, Yuancheng and Li, Lei and Ren, Shuhuai and Liu, Yuanxin and Ouyang, Kun and Wang, Lean and Li, Shicheng and Li, Sida and others},
  booktitle={ACM MM},
  year={2025}
}

@inproceedings{qian2025dispider,
  title={Dispider: Enabling video llms with active real-time interaction via disentangled perception, decision, and reaction},
  author={Qian, Rui and Ding, Shuangrui and Dong, Xiaoyi and Zhang, Pan and Zang, Yuhang and Cao, Yuhang and Lin, Dahua and Wang, Jiaqi},
  booktitle={CVPR},
  year={2025}
}

@article{zeng2025streamforest,
  title={StreamForest: Efficient Online Video Understanding with Persistent Event Memory},
  author={Zeng, Xiangyu and Qiu, Kefan and Zhang, Qingyu and Li, Xinhao and Wang, Jing and Li, Jiaxin and Yan, Ziang and Tian, Kun and Tian, Meng and Zhao, Xinhai and others},
  journal={NeurIPS},
  year={2025}
}

@article{tang2025video,
  title={Video understanding with large language models: A survey},
  author={Tang, Yunlong and Bi, Jing and Xu, Siting and Song, Luchuan and Liang, Susan and Wang, Teng and Zhang, Daoan and An, Jie and Lin, Jingyang and Zhu, Rongyi and others},
  journal={IEEE Transactions on Circuits and Systems for Video Technology},
  year={2025},
  publisher={IEEE}
}

@article{wu2024longvideobench,
    title={Longvideobench: A benchmark for long-context interleaved video-language understanding},
    author={Wu, Haoning and Li, Dongxu and Chen, Bei and Li, Junnan},
    journal={NeurIPS},
    year={2024}
}

@inproceedings{niu2025ovo,
  title={OVO-Bench: How Far is Your Video-LLMs from Real-World Online Video Understanding?},
  author={Niu, Junbo and Li, Yifei and Miao, Ziyang and Ge, Chunjiang and Zhou, Yuanhang and He, Qihao and Dong, Xiaoyi and Duan, Haodong and Ding, Shuangrui and Qian, Rui and others},
  booktitle={CVPR},
  pages={18902--18913},
  year={2025}
}

@inproceedings{huang2025online,
  title={Online Video Understanding: OVBench and VideoChat-Online},
  author={Huang, Zhenpeng and Li, Xinhao and Li, Jiaqi and Wang, Jing and Zeng, Xiangyu and Liang, Cheng and Wu, Tao and Chen, Xi and Li, Liang and Wang, Limin},
  booktitle={CVPR},
  pages={3328--3338},
  year={2025}
}

@article{lin2024streamingbench,
  title={Streamingbench: Assessing the gap for mllms to achieve streaming video understanding},
  author={Lin, Junming and Fang, Zheng and Chen, Chi and Wan, Zihao and Luo, Fuwen and Li, Peng and Liu, Yang and Sun, Maosong},
  journal={arXiv preprint arXiv:2411.03628},
  year={2024}
}

@article{wang2025internvideo,
  title={InternVideo2.5: Empowering Video MLLMs with Long and Rich Context Modeling},
  author={Wang, Yi and Li, Xinhao and Yan, Ziang and He, Yinan and Yu, Jiashuo and Zeng, Xiangyu and Wang, Chenting and Ma, Changlian and Huang, Haian and Gao, Jianfei and Dou, Min and Chen, Kai and Wang, Wenhai and Qiao, Yu and Wang, Yali and Wang, Limin},
  journal={arXiv preprint arXiv:2501.12386},
  year={2025}
}

@inproceedings{shen2024longvu,
  title={Longvu: Spatiotemporal adaptive compression for long video-language understanding},
  author={Shen, Xiaoqian and Xiong, Yunyang and Zhao, Changsheng and Wu, Lemeng and Chen, Jun and Zhu, Chenchen and Liu, Zechun and Xiao, Fanyi and Varadarajan, Balakrishnan and Bordes, Florian and others},
  booktitle={ICML},
  year={2025}
}

@article{fu2025vispeak,
  title={ViSpeak: Visual Instruction Feedback in Streaming Videos},
  author={Fu, Shenghao and Yang, Qize and Li, Yuan-Ming and Peng, Yi-Xing and Lin, Kun-Yu and Wei, Xihan and Hu, Jian-Fang and Xie, Xiaohua and Zheng, Wei-Shi},
  journal={ICCV},
  year={2025}
}

@article{chen2024internvl2,
  title={How far are we to gpt-4v? closing the gap to commercial multimodal models with open-source suites},
  author={Chen, Zhe and Wang, Weiyun and Tian, Hao and Ye, Shenglong and Gao, Zhangwei and Cui, Erfei and Tong, Wenwen and Hu, Kongzhi and Luo, Jiapeng and Ma, Zheng and others},
  journal={Science China Information Sciences},
  volume={67},
  number={12},
  pages={220101},
  year={2024},
  publisher={Springer}
}

@article{zhang2024longva,
  title={Long context transfer from language to vision},
  author={Zhang, Peiyuan and Zhang, Kaichen and Li, Bo and Zeng, Guangtao and Yang, Jingkang and Zhang, Yuanhan and Wang, Ziyue and Tan, Haoran and Li, Chunyuan and Liu, Ziwei},
  journal={arXiv preprint arXiv:2406.16852},
  year={2024}
}

@inproceedings{bolya2022tome,
  title={Token Merging: Your {ViT} but Faster},
  author={Bolya, Daniel and Fu, Cheng-Yang and Dai, Xiaoliang and Zhang, Peizhao and Feichtenhofer, Christoph and Hoffman, Judy},
  booktitle={ICLR},
  year={2023}
}

@inproceedings{di2025rekv,
  title={Streaming Video Question-Answering with In-context Video KV-Cache Retrieval},
  author={Di, Shangzhe and Yu, Zhelun and Zhang, Guanghao and Li, Haoyuan and Cheng, Hao and Li, Bolin and He, Wanggui and Shu, Fangxun and Jiang, Hao and others},
  booktitle={ICLR},
  year={2025}
}

@article{otsu1979threshold,
  author={Otsu, Nobuyuki},
  title={A threshold selection method from gray-level histograms},
  journal={IEEE Transactions on Systems, Man, and Cybernetics},
  year={1979},
  volume={9},
  number={1},
  pages={62--66}
}

@article{damonlpsg2023videollama,
  author={Zhang, Hang and Li, Xin and Bing, Lidong},
  title={Video-LLaMA: An Instruction-tuned Audio-Visual Language Model for Video Understanding},
  year={2023},
  journal={EMNLP},
  url={https://arxiv.org/abs/2306.02858}
}

@inproceedings{Maaz2023VideoChatGPT,
 author = {Maaz, Muhammad and Rasheed, Hanoona and Khan, Salman and Khan, Fahad Shahbaz},
 booktitle = {ACL},
 title = {Video-ChatGPT: Towards Detailed Video Understanding via Large Vision and Language Models},
 year = {2024}
}

@inproceedings{chen2024videollm,
  title={Videollm-online: Online video large language model for streaming video},
  author={Chen, Joya and Lv, Zhaoyang and Wu, Shiwei and Lin, Kevin Qinghong and Song, Chenan and Gao, Difei and Liu, Jia-Wei and Gao, Ziteng and Mao, Dongxing and Shou, Mike Zheng},
  booktitle={CVPR},
  year={2024}
}

@article{hurst2024gpt4o,
    title={Gpt-4o system card},
    author={Hurst, Aaron and Lerer, Adam and Goucher, Adam P and Perelman, Adam and Ramesh, Aditya and Clark, Aidan and Ostrow, AJ and Welihinda, Akila and Hayes, Alan and Radford, Alec and others},
    journal={arXiv:2410.21276},
    year={2024}
}

@article{team2024gemini,
    title={Gemini 1.5: Unlocking multimodal understanding across millions of tokens of context},
    author={Team, Gemini and Georgiev, Petko and Lei, Ving Ian and Burnell, Ryan and Bai, Libin and Gulati, Anmol and Tanzer, Garrett and Vincent, Damien and Pan, Zhufeng and Wang, Shibo and others},
    journal={arXiv:2403.05530},
    year={2024}
}

@article{qian2024videostreaming,
  title={Streaming long video understanding with large language models},
  author={Qian, Rui and Dong, Xiaoyi and Zhang, Pan and Zang, Yuhang and Ding, Shuangrui and Lin, Dahua and Wang, Jiaqi},
  journal={NeurIPS},
  volume={37},
  pages={119336--119360},
  year={2024}
}

@article{liu2026thinking,
  title={Thinking in Streaming Video},
  author={Liu, Zikang and Guo, Longteng and Li, Handong and Zhen, Ru and He, Xingjian and Ji, Ruyi and Ren, Xiaoming and Zhang, Yanhao and Lu, Haonan and Liu, Jing},
  journal={arXiv preprint arXiv:2603.12938},
  year={2026}
}

@article{xie2026fluxmem,
  title={FluxMem: Adaptive Hierarchical Memory for Streaming Video Understanding},
  author={Xie, Yiweng and He, Bo and Wang, Junke and Zheng, Xiangyu and Ye, Ziyi and Wu, Zuxuan},
  journal={arXiv preprint arXiv:2603.02096},
  year={2026}
}

@article{shen2026simple,
  title={A Simple Baseline for Streaming Video Understanding},
  author={Shen, Yujiao and Tian, Shulin and Yang, Jingkang and Liu, Ziwei},
  journal={arXiv preprint arXiv:2604.02317},
  year={2026}
}

@article{zhang2026weavetime,
  title={WeaveTime: Stream from Earlier Frames into Emergent Memory in VideoLLMs},
  author={Zhang, Yulin and Shi, Cheng and Yang, Sibei},
  journal={arXiv preprint arXiv:2602.22142},
  year={2026}
}

@inproceedings{khattab2020colbert,
  title={Colbert: Efficient and effective passage search via contextualized late interaction over bert},
  author={Khattab, Omar and Zaharia, Matei},
  booktitle={Proceedings of the 43rd International ACM SIGIR conference on research and development in Information Retrieval},
  pages={39--48},
  year={2020}
}

@article{zhang2026hermes,
  title={HERMES: KV Cache as Hierarchical Memory for Efficient Streaming Video Understanding},
  author={Zhang, Haowei and Yang, Shudong and Fu, Jinlan and Ng, See-Kiong and Qiu, Xipeng},
  journal={arXiv preprint arXiv:2601.14724},
  year={2026}
}

@article{fu2025contextnav,
  title={Contextnav: Towards agentic multimodal in-context learning},
  author={Fu, Honghao and Ouyang, Yuan and Chang, Kai-Wei and Wang, Yiwei and Huang, Zi and Cai, Yujun},
  journal={arXiv preprint arXiv:2510.04560},
  year={2025}
}

@article{mei2026gated,
  title={Gated Differentiable Working Memory for Long-Context Language Modeling},
  author={Mei, Lingrui and Liu, Shenghua and Wang, Yiwei and Ge, Yuyao and Bi, Baolong and Yao, Jiayu and Wan, Jun and Yin, Ziling and Guo, Jiafeng and Cheng, Xueqi},
  journal={arXiv preprint arXiv:2601.12906},
  year={2026}
}

@article{mei2025survey,
  title={A survey of context engineering for large language models},
  author={Mei, Lingrui and Yao, Jiayu and Ge, Yuyao and Wang, Yiwei and Bi, Baolong and Cai, Yujun and Liu, Jiazhi and Li, Mingyu and Li, Zhong-Zhi and Zhang, Duzhen and others},
  journal={arXiv preprint arXiv:2507.13334},
  year={2025}
}

@article{ge2025framemind,
  title={FrameMind: Frame-Interleaved Video Reasoning via Reinforcement Learning},
  author={Ge, Haonan and Wang, Yiwei and Chang, Kai-Wei and Wu, Hang and Cai, Yujun},
  journal={arXiv preprint arXiv:2509.24008},
  year={2025}
}

@article{wu2026camreasoner,
  title={CamReasoner: Reinforcing Camera Movement Understanding via Structured Spatial Reasoning},
  author={Wu, Hang and Cai, Yujun and Li, Zehao and Ge, Haonan and Sun, Bowen and Yuan, Junsong and Wang, Yiwei},
  journal={arXiv preprint arXiv:2602.00181},
  year={2026}
}

\clearpage
\appendix

\section{Detailed Problem Formulation and Analysis}
\label{supp:problem}

This section extends the streaming constraints introduced in the main paper, provides a taxonomy of visual memory representations, discusses the perception--memory trade-off, and clarifies evaluation protocols.

%% ----------------------------------------------------------
\subsection{Streaming Constraints: Extended Discussion}
%% ----------------------------------------------------------

\textbf{C1 --- No future frame access.}
The model may not access any frame $v_t$ with $t > T$ at any stage of processing, including visual encoding, memory construction, and token selection.

\textbf{C2 --- Query-agnostic encoding.}
\label{supp:c2}
Because $q$ is unknown during the encoding phase ($t < T$), all memory construction and token retention decisions must be made without conditioning on $q$.
This rules out query-aware compression strategies that use attention scores between visual tokens and the specific user query to guide token eviction.

We distinguish three levels of query dependence:
\begin{enumerate}[leftmargin=1.5em]
    \item \textbf{Query-dependent.} Compression is conditioned on the actual user query $q$. This violates C2.
    \item \textbf{Proxy-query-guided.} Compression uses a fixed set of generic pseudo-questions (e.g., ``What objects are visible?'') that are determined at system initialization and remain constant across all videos and queries. These carry no information about $q$ and serve only as content-agnostic importance priors. This is compliant with C2. StreamMem~\cite{yang2025streammem} adopts this approach, using chat template tokens as a generic proxy and showing empirically that it achieves comparable performance to using an explicit generic question (e.g., ``What is happening in the video?'').
    \item \textbf{Purely visual.} Compression relies exclusively on visual-level signals such as inter-frame cosine similarity. This is the strictest form of query-agnostic operation.
\end{enumerate}

\textbf{C3 --- Bounded memory.}
A compliant method must maintain a memory footprint bounded by a constant $B$ 
independent of video length.
Boundedness must hold for \emph{every} memory tier: if any tier grows without 
bound, the system cannot guarantee constant-memory operation.
Methods that store complete KV caches for all observed frames, such as 
ReKV~\cite{di2025rekv}, grow linearly and are treated as an oracle upper 
bound rather than a deployable solution~\cite{yang2025streammem}.
Among bounded-memory methods, LiveVLM~\cite{ning2025livevlm} compresses KVs 
via attention-based pruning and frame-wise merging, but discards earlier 
tokens when the memory upper bound is reached, risking forgetting of early 
content~\cite{yang2025streammem}; StreamMem~\cite{yang2025streammem} jointly 
re-compresses all stored KVs together with newly arriving ones at each step, 
maintaining a fixed-size memory throughout the video stream.
A well-designed system should make per-tier capacity explicit, so that the 
aggregate bound $B$ is a sum of known constants.

%% ----------------------------------------------------------
\subsection{Visual Memory: Retrieval vs.\ Replay}
%% ----------------------------------------------------------

Memory-based methods differ in the \textit{representation level} at which visual information is stored. This determines whether inference constitutes \textit{memory retrieval} (selecting pre-encoded representations) or \textit{video replay} (re-processing visual content through the encoding pipeline).

\textbf{(i) Raw frame retrieval and re-encoding.}
The most direct approach stores original frames and, at query time, retrieves relevant ones and passes them through the full vision encoder.
Although a retrieval step selects a subset, the retrieved content requires a complete encoding forward pass --- computationally identical to processing a shorter video from scratch. This is closer to \textit{selective replay}.

\textbf{(ii) KV-cache storage and retrieval.}
The predominant paradigm stores LLM key-value cache entries produced during encoding.
ReKV~\cite{di2025rekv} offloads KV features to RAM/disk and retrieves relevant entries at query time without re-invoking the visual encoder.
LiveVLM~\cite{ning2025livevlm} and StreamMem~\cite{yang2025streammem} adopt similar strategies.
Retrieval injects pre-encoded KV entries directly into decoding attention --- constituting memory retrieval.

\textbf{(iii) Hidden-state and compressed token memory.}
Flash-VStream~\cite{zhang2025flash} maintains a Flash Memory comprising 
a Context Synopsis Memory that aggregates temporal information via 
K-means clustering of encoded feature maps, and a Detail Augmentation 
Memory that stores high-resolution features from selected key frames.
These methods apply additional lossy compression that further distances stored representations from the original input.

\textbf{(iv) Text-level abstraction.}
Some methods generate captions and retrieve them via text similarity, discarding all sub-symbolic visual information.

\noindent We propose three criteria for \textit{retrieval} (as opposed to replay):
\textbf{(a)} \textit{One-pass encoding}: the visual encoder is invoked exactly once, with no re-invocation at query time;
\textbf{(b)} \textit{Irreversible transformation}: stored representations preclude reconstruction of the original visual input;
\textbf{(c)} \textit{Direct integration}: retrieved representations are injected directly into the LLM's decoding process without an additional encoding pass.
Methods in categories (ii)--(iv) satisfy all three criteria and constitute retrieval; category (i) does not.

\subsection{Query-Time Retrieval and Streaming Compliance}

A natural question is whether using the query $q$ at inference time
to select memory entries conflicts with the streaming constraints.
We note that C1 and C2 govern the \textit{encoding phase} ($t < T$):
memory must be constructed causally and without knowledge of $q$.
The \textit{inference phase}, triggered at $t = T$ when $q$ arrives,
operates over already-constructed memory and does not access any
frame beyond $T$. Since the memory content is entirely determined
before $q$ is known, conditioning retrieval on $q$ at inference
time does not introduce information leakage into the memory
construction process.

This two-phase structure, query-agnostic encoding followed by
query-conditioned retrieval, is the standard design in existing
streaming methods. ReKV~\cite{di2025rekv} encodes video with
sliding-window attention and stores all KV caches during encoding,
then retrieves query-relevant entries upon receiving a question.
LiveVLM~\cite{ning2025livevlm} continuously generates and compresses
video KVs during streaming, and selects query-relevant KVs when a
new question arrives. StreamMem~\cite{yang2025streammem} compresses
KV caches in a query-agnostic manner during encoding, and answers
questions over the compressed memory at inference time.
Flash-VStream~\cite{zhang2025flash} continuously updates its memory
in a frame handler process, while a separate question handler
reads from the memory to generate responses upon user queries.

%% ----------------------------------------------------------
\subsection{Pseudo-Streaming vs.\ True Streaming}
%% ----------------------------------------------------------

\textit{Pseudo-streaming} is the dominant evaluation protocol~\cite{di2025rekv,ning2025livevlm,yang2025streammem,zhang2025flash}: each QA pair is evaluated by truncating the video at query timestamp $T$ and feeding $\mathcal{V}_{[0,T]}$ to the model.
This correctly enforces C1 and C2, and is the standard of StreamingBench~\cite{lin2024streamingbench} and OVO-Bench~\cite{niu2025ovo}.
However, it does \textit{not} enforce strictly causal frame-by-frame encoding: the model may process all frames in $\mathcal{V}_{[0,T]}$ jointly, which would be impossible when frames arrive one by one in real time.

\textit{True streaming} requires strictly causal, frame-by-frame encoding with real-time latency constraints, as targeted by Dispider~\cite{qian2025dispider} and VideoLLM-online~\cite{chen2024videollm}.

We follow pseudo-streaming, consistent with existing methods and benchmarks, enabling direct comparison under identical conditions.
We acknowledge it does not penalize encoding latency, and extending to true streaming evaluation remains important future work.

\section{Pseudo Question}
\label{sup:pseudo}
\begin{lstlisting}[style=mypython, caption={Pseudo-question bank}, label={lst:pseudo-q}]
_PSEUDO_QUESTIONS = (
    "What objects are visible in the scene?",
    "How many items or people can be seen?",
    "What actions or events are happening?",
    "What has changed in the scene?",
    "Describe the spatial arrangement of objects.",
)
\end{lstlisting}

\section{Social Impact}
Our method improves the efficiency of online streaming video understanding by reducing GPU memory usage in a training-free manner, which can lower the deployment cost of real-time video assistants and benefit applications such as live captioning and assistive perception. As with general-purpose video understanding models, potential misuse in surveillance scenarios exists, but our work does not introduce new capabilities beyond those of the backbone model and poses no additional societal risks.

\end{document}